\documentclass[conference]{IEEEtran}
\IEEEoverridecommandlockouts
\usepackage{cite}
\usepackage{amsmath,amssymb,amsfonts}
\usepackage{graphicx}
\usepackage{textcomp}
\usepackage{xcolor}
\usepackage{siunitx}
\usepackage{tabularx}
\usepackage[colorinlistoftodos]{todonotes}
\usepackage[acronym]{glossaries}
\usepackage{algorithm}
\usepackage{algpseudocode}
\usepackage{soul}
\usepackage{multirow}
\usepackage{multicol,caption}
\usepackage{lipsum}
\usepackage{subcaption}
\usepackage{comment}
\usepackage{booktabs}
\usepackage{siunitx}

\usepackage{datetime}
\usepackage{enumerate}
\usepackage{booktabs}
\usepackage{flushend}
\usepackage{caption}
\usepackage{pdflscape}
\usepackage{rotating}
\usepackage{wrapfig}
\usepackage{tabularx}
\usepackage{enumitem}
\usepackage[hyphens]{url}
\usepackage{pifont}%
\newcommand{\cmark}{\ding{51}}%
\newcommand{\xmark}{\ding{55}}%

\definecolor{hlgreen}{HTML}{d9ead3} 
\definecolor{hlblue}{HTML}{c9daf8} 

\newcommand{\hlgreen}[1]{%
  \begingroup
  \sethlcolor{hlgreen!40}%
  \hl{#1}%
  \endgroup
}

\newcommand{\hlblue}[1]{%
  \begingroup
  \sethlcolor{hlblue!40}%
  \hl{#1}%
  \endgroup
}

\def\BibTeX{{\rm B\kern-.05em{\sc i\kern-.025em b}\kern-.08em
    T\kern-.1667em\lower.7ex\hbox{E}\kern-.125emX}}
\begin{document}

\title{\emph{ROS Help Desk}:
GenAI Powered, User‑Centric Framework for ROS Error Diagnosis and Debugging
\\\thanks{This work was funded by CSIRO’s Data61 Science Digital}
}

\author{\IEEEauthorblockN{Kavindie Katuwandeniya$^{\dagger}$}
\IEEEauthorblockA{\textit{CSIRO Robotics} \\
Clayton, Melbourne, Australia \\
kavi.katuwandeniya@csiro.au}
\and
\IEEEauthorblockN{Samith Rajapaksha Jayasekara Widhanapathirana $^{\dagger}$}
\IEEEauthorblockA{\textit{CSIRO Robotics} \\
Clayton, Melbourne, Australia \\
samith.ashan@data61.csiro.au}
\thanks{$^{\dagger}$ These authors contributed equally to this work.}
}
\maketitle

\begin{abstract}
As the robotics systems increasingly integrate into daily life, from smart home assistants to the new-wave of industrial automation systems (Industry 4.0), there’s an increasing need to bridge the gap between complex robotic systems and everyday users. 
The Robot Operating System (ROS) is a flexible framework often utilised in writing robot software, providing tools and libraries for building complex robotic systems.
However, ROS's distributed architecture and technical messaging system create barriers for understanding robot status and diagnosing errors. This gap can lead to extended maintenance downtimes, as users with limited ROS knowledge may struggle to quickly diagnose and resolve system issues. Moreover, this deficit in expertise often delays proactive maintenance and troubleshooting, further increasing the frequency and duration of system interruptions. 
\emph{ROS Help Desk}  provides intuitive error explanations and debugging support, dynamically customized to users of varying expertise levels. 
It features user-centric debugging tools that simplify error diagnosis, implements proactive error detection capabilities to reduce downtime, and integrates multimodal data processing for comprehensive system state understanding across multi-sensor data (e.g., lidar, RGB). 
Testing qualitatively and quantitatively with artificially induced errors demonstrates the system's ability to proactively and accurately diagnose problems, ultimately reducing maintenance time and fostering more effective human-robot collaboration.
\end{abstract}

\section{Introduction}
Robotic systems are expanding their presence in diverse environments due to advancements in Generative Artificial Intelligence (GenAI) capabilities. However, these systems remain susceptible to failures. Therefore, a user-centric design emphasizing transparent interaction is essential. 
This allows for adaptive responses to system anomalies, minimizing disruptions and facilitating seamless user involvement in ensuring operational continuity. 
Providing clear explanations is crucial for establishing trust and comprehension between people and their robotic counterparts. 
Explainability relates to interpretability and transparency \cite{scheltingaexplaining} where the goal is to improve user’s understanding of the robot through clear and truthful explanations that align with the robot’s logic. 

\begin{figure}[ht]
    \centering
    \includegraphics[width=\linewidth]{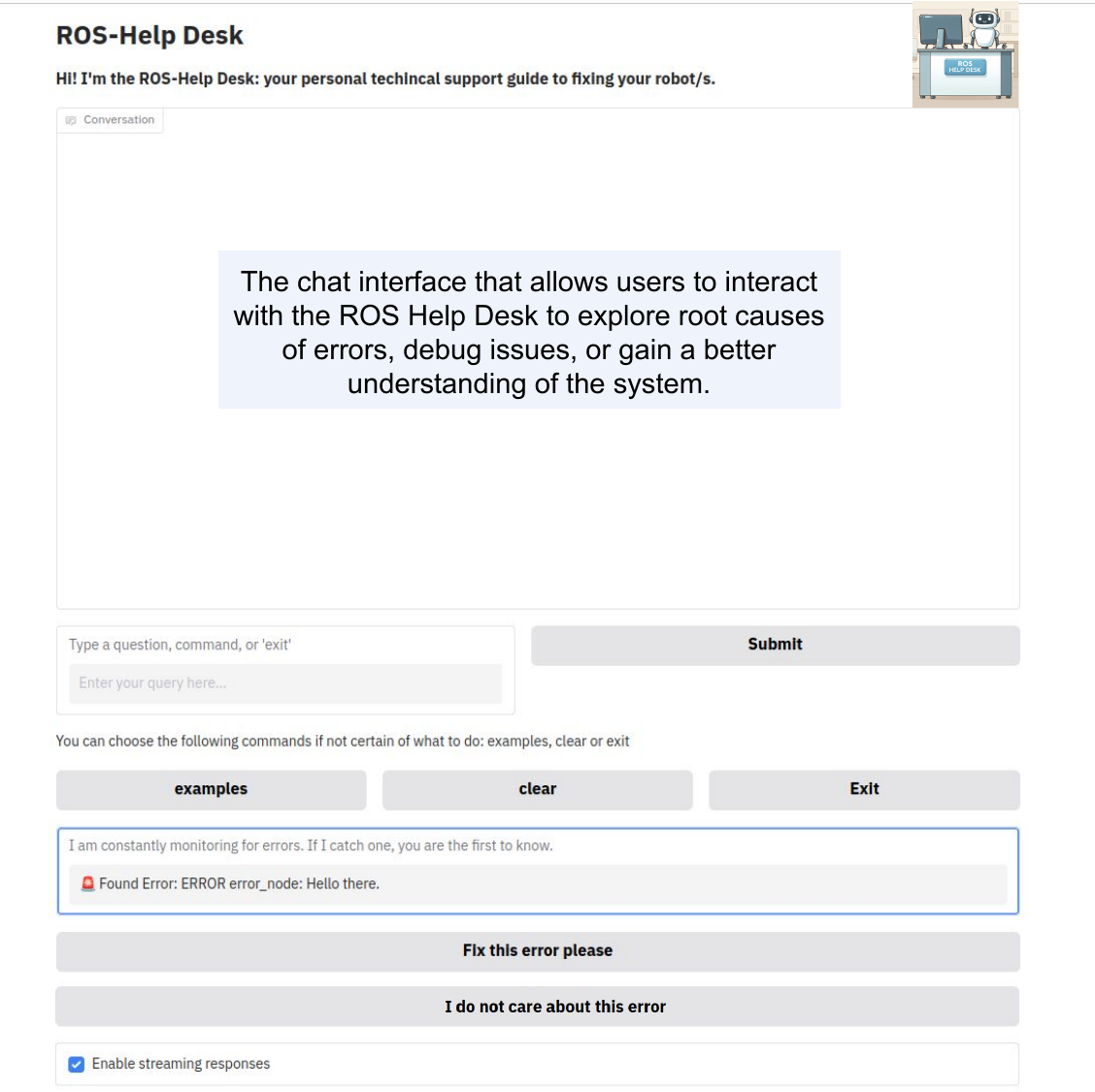}
    \caption{User interface of the \emph{ROS Help Desk} chat system, enabling users to diagnose errors, debug issues, and explore system behaviour through interactive conversation. Real-time error diagnosis messages appear at the bottom of the screen, allowing users to either explore or ignore them. If explored, the message is added to the chat interface for further querying.} %
    \label{fig:interface}
\end{figure}

A significant portion of robotic systems relies on the open-source framework: Robot Operating System (ROS)~\cite{quigley2009ros}, with an estimated 55\% of commercial robots shipped in 2024—over 915,000 units—containing at least one ROS package~\cite{Koh2023}.  
Within ROS, data exchange occurs between nodes via topics, a publish-subscribe messaging system that enables asynchronous and decoupled communication crucial for distributed robotic architectures. 
Detailed operational information is captured in textual log messages published to the 
\emph{/rosout} topic. This log data offers a standardized view of internal robot processes, facilitating cross-system comparisons. 
However, traditional approaches to programming and controlling these robots often necessitate specialized technical expertise, rendering debugging complex and inaccessible for many operators. This expertise gap creates significant barriers to effective troubleshooting and impedes the broader adoption of robotic technologies. 
Importantly, not all errors within robotic systems are reliably captured via log messages, necessitating alternative diagnostic methods. 
Furthermore, current systems often lack the capability for proactive error detection, leading to retroactive troubleshooting and increased downtime. 
To address both the challenges of accessibility and the limitations of log-based debugging, the \emph{ROS Help Desk} system is developed to provide a user-friendly medium that empowers operators with diverse expertise levels to not only detect errors but also contribute to the debugging process within the robotic domain.

To create user-friendly, interactive explanations, we explore the use of Large Language Models (LLMs) for analysing the substantial textual data found ROS log messages and generating clear responses about robot status and actions. 
Despite their potential, LLMs may produce factually incorrect outputs~\cite{ji2023survey} and must adapt to users’ expertise levels.
We introduce \emph{ROS Help Desk}, a framework supporting robotic debugging through proactive error detection, multimodal data integration (e.g., vision, lidar), and user-tailored explanations. 
Unlike standard multimodal LLMs, \emph{ROS Help Desk} contextualizes sensor data within ROS operational semantics via custom preprocessing pipelines, enabling the identification of correlations between sensor anomalies and system failures. The framework also includes tools for code review, retrieval-augmented generation (RAG)~\cite{lewis2020retrieval} with an evolving error database, and a graphical user interface. Together, these components help bridge the gap between users and complex ROS-based robotic systems. 

\section{Related Work}
\subsubsection{Robots Following Instructions}
The use of natural language instructions to control robots has seen significant growth, with research exploring various approaches to bridge the gap between human intent and robotic action. 
More recently, the advent of LLMs has revolutionized this field, enabling robots to interpret and execute more complex and ambiguous instructions. 
Tellex~et~al.~\cite{tellex2020robots} reviewed the use of natural language in robotics, finding that most research focuses on robots following human instructions~\cite{lynch2023interactive}. 
Existing research exhibits a notable gap in the exploration of robust, robot-to-human communication or two-way communication. Furthermore, the specific application of such communication for robotic system diagnostics and repair remains largely unaddressed in current literature. 

\subsubsection{ROS-based Developments}
Developing frameworks that translate natural language to ROS interfaces and vice-versa is crucial for enabling users to command and understand robots in a natural way without requiring specialized programming knowledge, thereby broadening accessibility and efficiency in robotic applications.

ROSGPT~\cite{koubaa2025next} claims to be the first to combine ROS with an LLM (ChatGPT), bridging the gap between natural language understanding and ROS.  They use prompt engineering and ontology development to translate human language commands into structured robotic instructions. A proof-of-concept demonstrates this with spatial navigation, and quantitative evaluations across various robots and LLMs, showing promising results. 
OperateLLM~\cite{raja2024operatellm}, employs the ReAct~\cite{yao2023react} method of prompting (reasoning and acting) to interstage ROS2 with LLMs. By using the DeepSeek Coder v2`%
as the core LLM and the rclpy Python API %
for ROS2, OperateLLM facilitates the creation of basic ROS2 components like nodes and publishers, as well as more complex tasks. 
ROS-LLM~\cite{mower2024ros} enables non-experts to program robots intuitively using natural language prompts and contextual data from ROS. By integrating LLMs, ROS-LLM allows users to specify tasks through a chat interface. The system extracts behaviours from LLM outputs and executes ROS actions/services, supporting sequence execution, behaviour trees, and state machines. 

Robotic Operating System Agent (ROSA)~\cite{royce2024enabling} pushes these advances further by following a Cognitive Architecture for Language Agents~\cite{sumers2023cognitive}.  Their overall architecture consists of four primary components: the action space, a set of memory modules, internal logic for decision making, and the LLM.  ROSA's design is modular and extensible, allowing for integration with both ROS1 and ROS2, and incorporates safety mechanisms.  It enhances human-robot interaction by making complex robotic systems more accessible, supporting multi-modal capabilities like speech and visual perception. 

Current efforts in natural language to ROS interfaces primarily concentrate on direct code generation for task execution. However, essential functionalities such as providing explanations for robot actions, proactive error detection and correction mechanisms, and offering debugging tools are significant areas lacking substantial research where the later two is the focus of this work. 
Additionally, the expertise level of the user is not taken into consideration in any of these work. 

\subsubsection{ROS-based Failure Detection and Debugging}

Traditional ROS debugging relies on a suite of tools, including command-line utilities like 
\textit{ros2 doctor} for system health checks, graphical rqt tools~\cite{rqt} for visualization and data plotting, and rviz~\cite{rviz} for 3D sensor data inspection. While powerful for experienced developers, these tools often demand significant technical expertise, are primarily retrospective in nature, and lack user-level debugging support. 

Scheltinga and Pek~\cite{scheltingaexplaining}  propose a framework to finetune LLMs to explain autonomous robot navigation failures as a crucial step for building user trust. Unlike traditional debugging, this work focuses on interpreting and communicating the  reasons behind a robot's limitations, not just fixing code. By analyzing ROS log data, the authors aim to provide human-understandable explanations of why a robot failed, marking a shift towards enhancing robot explainability for broader user understanding.

The \emph{ROS Help Desk} framework focuses on extending this further, by integrating proactive error detection, adapting the explanation per user and also providing debugging support giving more autonomy to non-technical users, allowing wide-adaptation of robotics. 

\begin{figure}[ht]
    \centering
    \includegraphics[width=\linewidth]{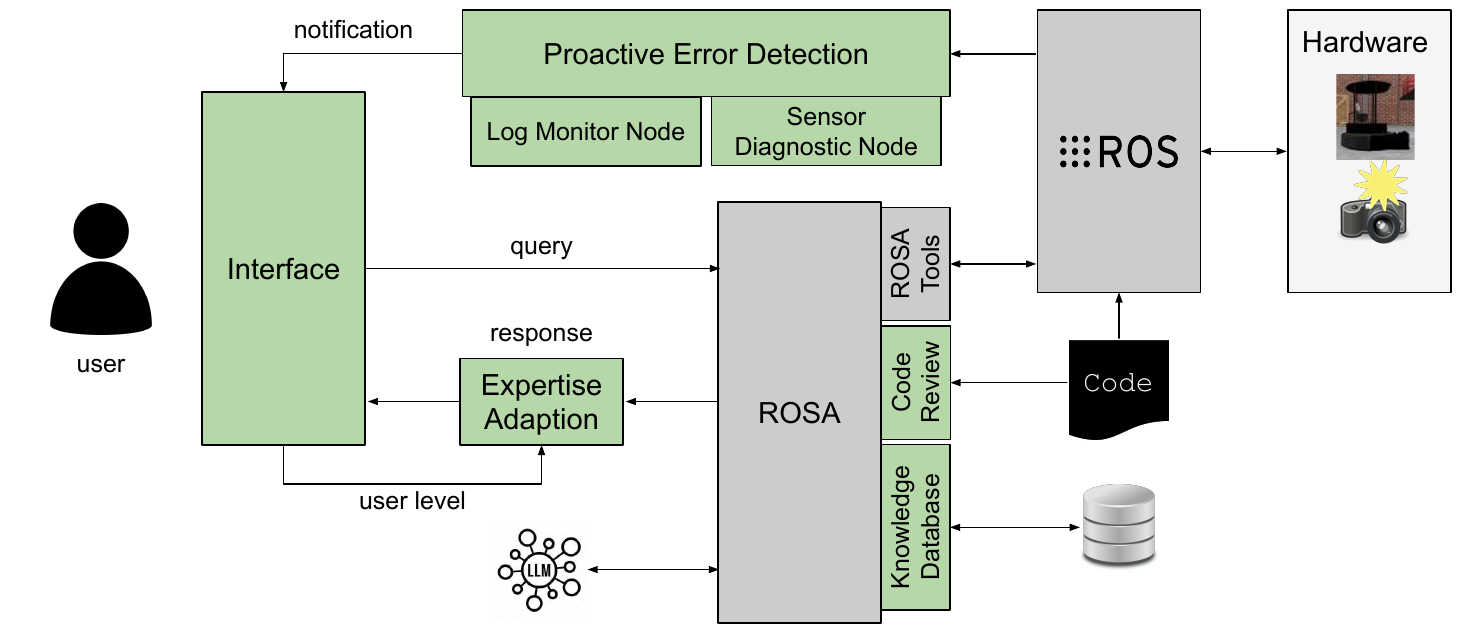}
    \caption{ROSA Help Desk Architecture}
    \label{fig:architecture}
\end{figure}

\section{System Architecture}
The architecture of the \emph{ROS Help Desk} is shown in Fig.~\ref{fig:architecture} which builds upon the foundational work of Robotic Operating System Agent (ROSA)~\cite{royce2024enabling}, extending its capabilities to provide a more robust and user-centric debugging experience within ROS-based robotic systems. It augments LLM-based reasoning with real-time monitoring capabilities, multimodal sensor diagnostics, adaptive user modelling, and a dynamic knowledge base of error patterns and resolutions. 
The system architecture is designed to provide personalized assistance across varying levels of technical expertise, facilitating efficient detection, interpretation, and resolution of issues during robotic system operation.

\subsubsection{Real-time Error Diagnosis and Multimodal Contextualization}
\emph{ROS Help Desk} employs a dual-pronged approach to error detection that addresses both explicit and implicit system issues.

A dedicated ROS node (\textbf{Log Monitor Node}) continuously monitors the $/rosout$ topic, parsing log messages to detect exceptions, and fatal errors. Upon detection, these issues are automatically forwarded to the LLM agent for interpretation and resolution generation. This provides immediate insights into common software-related issues such as communication failures, parameter misconfigurations, and runtime exceptions.

Recognizing that not all error sources manifest directly in log messages, the framework integrates real-time sensor data analysis as a crucial step toward proactive error detection. 
A specialized diagnostic node monitors multimodal sensor data streams (\textbf{Sensor Diagnostic Node})—for this work, from cameras and lidar sensors—to identify anomalies such as missing frames, blank images, or invalid point cloud returns that might precede or exist independently of explicit ROS errors. For instance, unexpected deviations in lidar readings could indicate environmental changes affecting navigation, while unusual patterns in image data might suggest sensor malfunctions effecting object recognition. 

By analyzing these sensor streams in parallel with log data, the framework builds a comprehensive understanding of the robot's operational state and can identify subtle indicators of potential problems before they escalate into critical system failures. This integration of multimodal data sources significantly enhances the system's error detection capabilities beyond traditional log-based monitoring.

\subsubsection{User Expertise Adaption}
To deliver personalized assistance, the system incorporates an initial interaction phase where users self-report (explicit) their technical background (beginner, intermediate, expert). 
This assessment, combined with interaction history analysis (implicit), allows the system to dynamically adapt its communication style and technical depth. %

\subsubsection{Code Review Module}
When prompted by users or when error tracebacks reference specific files, this module analyzes ROS node scripts to identify potential syntax errors, logic flaws, or configuration issues. A LangChain-powered tool parses source code to extract relevant functions, variables, and configuration parameters.

\subsubsection{Evolving Knowledge Database} 
ROSA is built using ReAct (Reasoning and Acting)~\cite{yao2023react} agents, which follow a loop of reasoning, taking actions via tools (Python functions in ROSA), and observing outcomes to inform the next step. The ReAct agents are extended with retrieval capabilities, effectively combining ReAct with RAG~\cite{lewis2020retrieval}.

The system maintains a continuously updated repository of previously encountered errors and their resolutions. This database serves as a critical reference during the LLM's reasoning process, enabling efficient retrieval of relevant historical cases and avoiding redundant diagnostics.

To overcome the need for specialized databases with ROS errors, a two-stage retrieval process is implemented: first applying keyword matching to filter potentially relevant cases, then employing vector embedding matching using Microsoft CodeBERT for more precise semantic similarity. 
The RAG+LLM system leverages this efficient filtering mechanism while utilizing the LLM's reasoning to interpret and adapt historical solutions to current contexts—providing both retrieval speed and generative flexibility. Following successful resolution sessions, new error-fix pairs are appended to the database, supporting long-term adaptability and performance improvement of the system.

\subsubsection{User Interface}
The Gradio-based~\cite{abid2019gradio} graphical interface features interactive real-time error notifications, and natural language query input. When errors are detected, users receive a pop-up notifications detailing the issue and a button to click if they want the issue fixed. If it is the case, the system proposes and executes actions to fix the error - and the final result is shown to the user indicating the error is fixed or inquiring if users would want to debug further. The interface design prioritizes clarity and accessibility, with error messages presented alongside relevant contextual information and suggested resolutions tailored to the user's expertise level.

This comprehensive methodology establishes \emph{ROS Help Desk} as an intelligent debugging assistant capable of providing contextually relevant, personalized support for users interacting with complex ROS-based robotic systems.

\section{Evaluation}
The effectiveness of \emph{ROS Help Desk} was evaluated for its ability to detect, diagnose, and resolve common ROS2 errors through quantitative and qualitative assessments, utilizing a comprehensive fault injection framework.
Quantitatively, we examine three key dimensions: (1) proactive error detection capabilities compared against the ROSA baseline, (2) debugging accuracy and efficiency by evaluating system-generated solutions against expert-defined guidelines across multiple trials, and (3) expert evaluation where experienced ROS users rate the system. 
Qualitatively, we demonstrate the system's ability to adapt explanations to different user expertise levels through representative examples of system responses.

\subsection{Simulation Environment and Fault Injection}
\begin{figure}[tbp]
  \centering
  \begin{subfigure}[b]{0.15\textwidth}
    \centering
    \includegraphics[width=\linewidth]{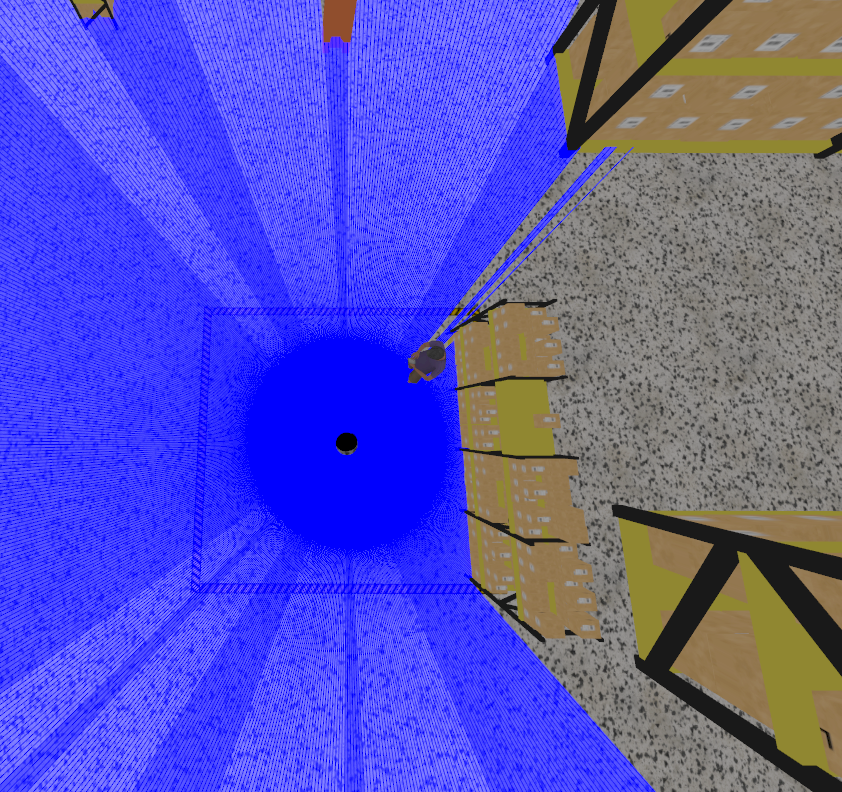}
    \caption{}
    \label{fig: lidar_error_gazebo}
  \end{subfigure}%
  \hfill
  \begin{subfigure}[b]{0.15\textwidth}
    \centering
    \includegraphics[width=\linewidth]{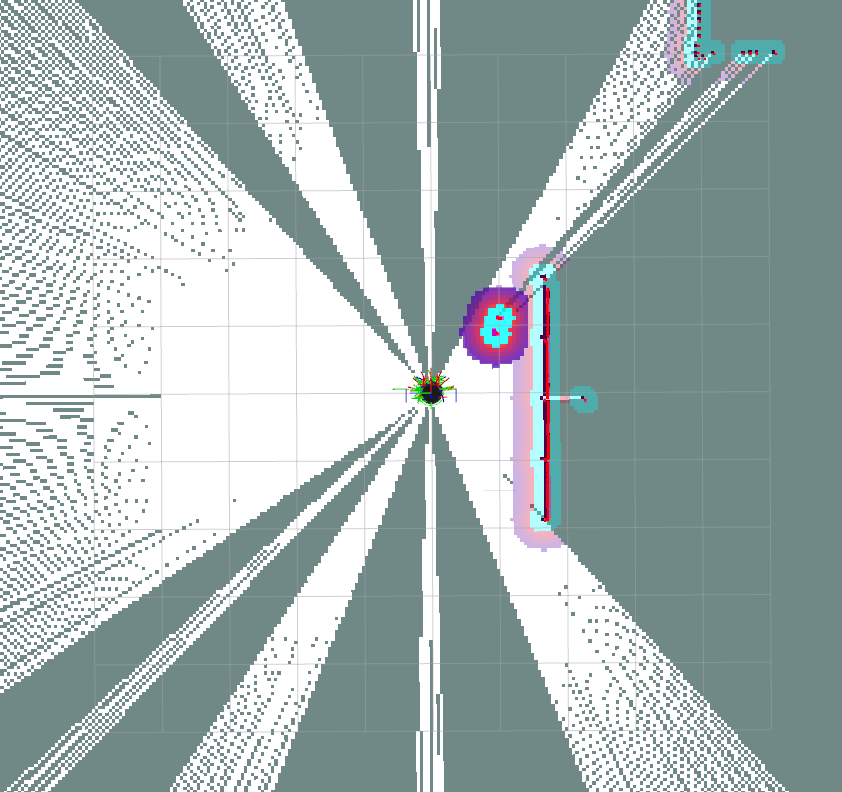}
    \caption{}
    \label{fig: Normal_rviz}
  \end{subfigure}%
  \hfill
  \begin{subfigure}[b]{0.15\textwidth}
    \centering
    \includegraphics[width=\linewidth]{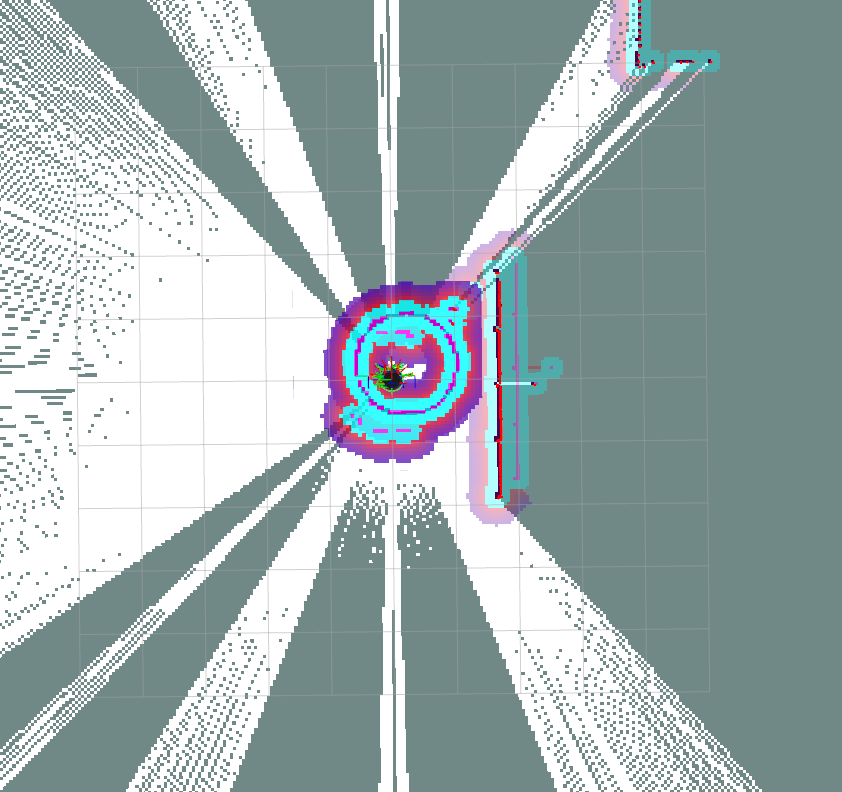}
    \caption{}
    \label{fig: lidar_error_rviz}
  \end{subfigure}
  \caption{Visualization of `corrupt' fault injection with lidar data. \emph{(left)} Lidar rays in Gazebo simulation environment.  \emph{(middle)} Normal condition: Lidar rays in Rviz, without fault injection on lidar data. \emph{(right)} Faulty condition: Lidar rays in Rviz, with corrupted lidar data.}
  \label{fig:lidar_error}
  \vspace{-2mm}
\end{figure}

As the first step, a controlled simulation environment was developed using Gazebo~\cite{koenig2004design} with a TurtleBot3 robot configured to perform navigation tasks in an indoor environment. Fig.~\ref{fig: lidar_error_gazebo} shows the robot in the environment. 

Then, a fault injection framework was developed to evaluate the robustness and effectiveness of \emph{ROS Help Desk} by simulating a range of error conditions within the ROS2-based system. The framework  artificially triggers faults in a controlled manner, and faults are injected by  simulating sensor data corruption through configuration of topic publishers and subscribers. 
As an initial step in the evaluation, the focus was on the following commonly observed ROS errors: Sensor Faults (sensor noise, sensor bias and other data corruption), Communication Faults (message loss and message delay), and Node Crashes.

The framework follows the methods outlined in \cite{Hsiao2021MAVFIAE},  utilizing a YAML file-based error configuration system to define the fault space. The configuration parameters include fault injected topic type (lidar or camera sensor), ROS message type, input and output topics, error type (``corrupted", ``drop", or ``delay"), error value (magnitude), and error frequency. This configuration approach enables systematic and reproducible fault injection across various components of the ROS2 system.

\begin{figure}[htb]
    \centering
    \includegraphics[width=\linewidth]{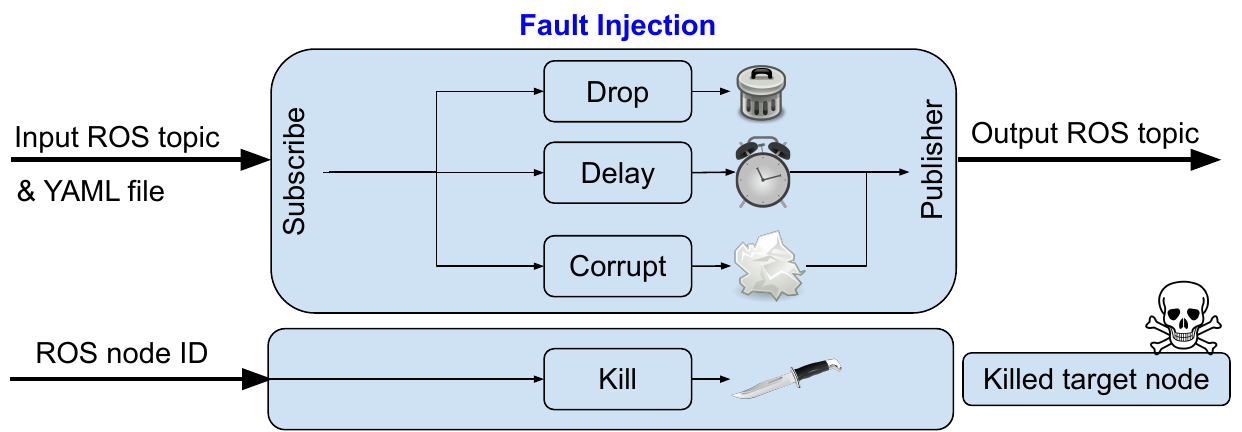}
    \caption{Fault Injection Framework Architecture. The framework intercepts communication between ROS nodes by remapping sensor topics through a fault injection node. %
    Error types (message corruption, delay, or drop) are applied to the sensor data stream before being published to the output topic. %
    Nodes are killed based on the given node ID. 
    }
    \label{fig:fault_injection}
    \vspace{-2mm}
\end{figure}

The fault injection process employs a custom ROS node that intercepts the normal message flow as shown in Fig.~\ref{fig:fault_injection}. Upon initialization, the node configures parameters based on the YAML specifications, establishes appropriate subscribers and publishers for the selected sensor types, and processes incoming messages according to the defined error parameters. When triggered by the probabilistic error frequency check, the node applies the specified fault type—replacing data with erroneous values for corruption faults, suspending publication for delay faults, or withholding messages entirely for drop faults. 
Fig.~\ref{fig:lidar_error} illustrate a lidar fault injection scenario, which is visualized in rviz, where the lidar rays are visualized, with the Fig.~\ref{fig: lidar_error_gazebo} showing the baseline simulation. The Fig.~\ref{fig: Normal_rviz} presents normal, unaltered lidar data, while the Fig.~\ref{fig: lidar_error_rviz} reveals the distortions introduced by fault injection.

Additionally, Node Crash faults are induced via a script which forcefully terminates the ROS node from its node ID.

\subsection{Quantitative Evaluation}
The framework is evaluated in two major aspects: proactive error detection, and debugging accuracy and efficiency. 

\begin{table*}[htb]
    \centering
    \begin{tabular}{l*2c*9{S[table-format=3\%]}}
        \toprule
                        &\multicolumn{2}{c}{Error detection}
                                        &\multicolumn{9}{c}{Debugging evaluation} \\
                        \cmidrule(r){2-3} \cmidrule(l){4-12}
                        & ROSA  & Ours  &\multicolumn{9}{c}{Assessment criteria} \\
                                          \cmidrule(lr){4-11}
        Fault Category  & (queried) & (proactive)
                                        &  {A}  &  {B}  &  {C}  &  {D}  &  {E}  &  {F}  &  {G}  &  {H}  & {Average} \\\midrule
        Lidar Drop      &\xmark &\cmark &  80\% &  80\% &  80\% &  66\% &  54\% &  48\% &  68\% &  34\% &  64\%     \\
        Lidar Delay     &\xmark &\cmark &  80\% & 100\% &  40\% &  70\% &  66\% &  50\% &  84\% &  40\% &  66\%     \\
        Lidar Corrupt   &\cmark &\cmark &  60\% & 100\% & 100\% &  72\% &  66\% &  58\% &  78\% &  36\% &  71\%     \\\addlinespace
        Image Drop      &\xmark &\cmark & 100\% & 100\% & 100\% &  74\% &  70\% &  60\% &  80\% &   0\% &  73\%     \\
        Image Delay     &\xmark &\cmark &  80\% & 100\% & 100\% &  70\% &  66\% &  58\% &  78\% &   0\% &  69\%     \\
        Image Corrupt   &\cmark &\cmark &  80\% & 100\% & 100\% &  68\% &  60\% &  44\% &  78\% &   0\% &  66\%     \\\addlinespace
        Node Crash      &\xmark &\cmark & 100\% & {---} & 100\% &  66\% &  44\% &  42\% &  74\% &  14\% &  63\%     \\\midrule
        \textbf{Average}&  29\% & 100\% &  83\% &  97\% &  89\% &  69\% &  61\% &  51\% &  77\% &  18\% &  68\%     \\
        \bottomrule
    \end{tabular}
    \caption{Error identification accuracy (ROSA vs \emph{ROS Help Desk}) and evaluation of debug reports across different fault categories and assessment criteria (see Sec.~\ref{sec:criteria}).}
    \vspace{-2mm}
    \label{tab:combined_error_evaluation}
\end{table*}

\subsubsection{Proactive error detection}
The system's ability to detect different types of errors is assessed by checking whether the correct error appears in the interface. 
ROSA is used as a comparison baseline. Since ROSA does not have the capability to proactively diagnose errors, it is prompted with `Is there any error in image/lidar data', followed by `What is the anomaly in the image/lidar data' if it was successful in identifying an anomaly but did not elaborate further.
If the anomaly is not successfully identified to be either delayed, dropped, or corrupt,
the system is considered as unable to identify the error even when queried.  
The results are given in Tab.~\ref{tab:combined_error_evaluation}, columns 2 and 3.
Across all injected error types, the \emph{ROS Help Desk} successfully identified the correct error $100\%$ of the time.
Even after prompting, ROSA was unable to find the injected communication errors or node crash errors, leading to a total accuracy of $29\%$.

The effectiveness of the proactive error detection node comes not only from the ability to detect but also to meaningfully describe the fault. This description drastically narrows the error search space; without this, ROSA is left with no option but to try a random selection of diagnostic tools until it ultimately gives up.
A common tool used by ROSA was `ros2 topic echo', which displays the message in a textual format. From this output, the ROSA system is able to detect the corruption fault modes by simply observing repeated values. However, it should be noted that this is possible in part due to the simplistic corruption method used; more complicated corruption artefacts would equivalently require more complicated analysis to detect.

\subsubsection{Debugging accuracy and efficiency}

To objectively evaluate the system's output quality, 
the debugging solutions provided by \emph{ROS Help Desk} are assessed against expert-defined guidelines.
For each error type, debugging outputs are generated and analysed five times to account for any variability in the agent's reasoning. The debugging report is evaluated as per the following criterion:
\begin{enumerate}[label=\Alph*.]
\label{sec:criteria}
    \item Recognizes the relevant node,    
    \item Recognizes the relevant topic,
    \item Identifies the error type (drop, delay, etc.),
    \item Analyses the error with hypotheses of the cause,
    \item Performs diagnostic checks,
    \item Presents diagnostic results validating a hypothesis,
    \item Recommends further actions,
    \item Identifies the true cause (intentional fault injection).
\end{enumerate}
Criteria~A-C are easily assessed and scored by assigning boolean pass-fail values.
Criteria~D-H are more open-ended and intrinsically subjective.
To produce objective assessments, a different LLM (Claude) is employed to score the reports against the criteria, on a scale from 0-10.
The expert guidelines are elaborated to include statements such as `The relevant node is recognised to be \{node\_name\}', `The relevant topic is recognised to be \{topic\_name\}', `Bonus points if the issue is correctly identified as \{cause\}'.
It is worth noting that the LLM's assessment aligns reasonably well with the author's human assessment. 
This approach allowed us to quantitatively measure solution quality aspects. %
The results are given in Tab.~\ref{tab:combined_error_evaluation}, columns 4 to 12.

The identification of relevant nodes, topics, and error types (criteria A, B, and C) demonstrated high accuracy rates. This performance can be attributed to the effectiveness of real-time error diagnosis mechanisms, which generate precise error messages that effectively guide the LLM agent toward appropriate diagnostic pathways.

Criteria D through G each exhibited accuracy rates exceeding $50\%$, indicating substantial capability in the agent's reasoning processes and follow-through mechanisms. The implementation of additional diagnostic tools may potentially enhance performance metrics for these criteria.
Criterion H, which involves identification of intentionally injected faults, was not established as a primary evaluation metric due to the testing mechanism's dissimilarity to real-world applications. Nevertheless, through code review capabilities, the agent successfully identified intentional fault injections  $18\%$ of the time.

It is worth noting that a number of diagnostic sessions were compromised by spurious tool execution failures within the ROSA system. These technical interruptions diminished the agent's diagnostic capabilities, resulting in incomplete debug reports. System reliability could be enhanced through implementation of more robust tools (such as accommodating syntax variations: ``/node\_name" versus ``node\_name"), thereby improving overall diagnostic effectiveness and report completeness.

\subsubsection{Expert Evaluation}
Three daily ROS users %
were enlisted to use and rate the \emph{ROS Help Desk} system in helping find an error (corrupt images) and debugging solutions, in terms of five aspects: clarity, time efficiency, solution applicability, technical appropriateness,and overall effectiveness. The system achieved a mean score of 4 out of 5 across all evaluation criteria, representing a promising foundation for continued development. These encouraging results justify the implementation of a comprehensive user study to further validate performance and inform subsequent enhancements.

\begin{figure}[ht]
    \centering
    \includegraphics[width=\linewidth]{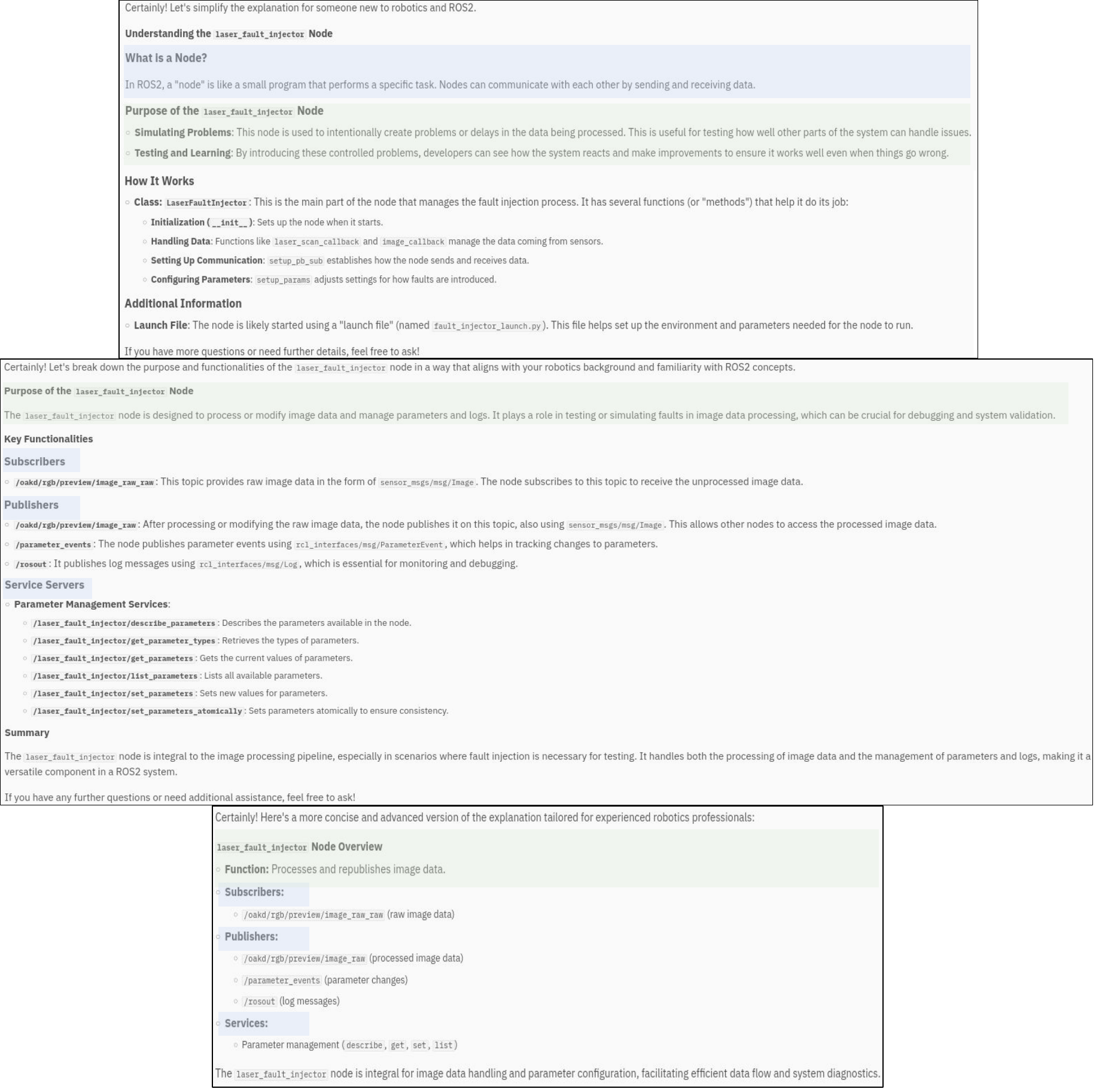}
    \caption{Qualitative evaluation of the ability of ROSA Help Desk to adapt to different expertise levels. 
    From top to bottom the expertise levels are beginner, intermediate and expert. 
    The query is ``Tell me about the purpose of the laser\_fault\_injector node''. \hlblue{Blue} highlights the explanation or lack of ROS-related terms and \hlgreen{green} highlights the explanation of the purpose of the node. 
}
    \vspace{-2mm}
    \label{fig:expertise_adaption}
\end{figure}

\subsection{Qualitative Evaluation}
The ability of \emph{ROS Help Desk} to adapt to different user expertise is shown qualitatively in Fig.~\ref{fig:expertise_adaption} which shows the answer provided to the query `Tell me about the purpose of the laser\_fault\_injector node'. Moving from beginner to intermediate it is visible that the explanation of the common ROS terms like node is dropped and from intermediate to expert, the description is very compact and technical. More details about the node such as its publishers, subscribers and services are given to the intermediate and expert users while that information omitted from beginners.  

\section{Discussion and Future Work}
We propose \emph{ROS Help Desk}: an accessible interface enabling operators of all expertise levels to proactively detect errors and participate in debugging processes within robotic environments. The system was evaluated using a fault injection framework covering  various error scenarios.
While our fault injection methodology captures common ROS2 error patterns, it cannot encompass the full spectrum of failures encountered in real-world robotics deployments. Moreover, this artificial setting with a purpose-built fault injection node cannot perfectly replicate naturally occurring errors in terms of their manifestation patterns and cascading side effects. As such, better evaluation settings need to be explored. 
\emph{ROS Help Desk} provides real-time error diagnosis. This proactive methodology offers preventive benefits but operates on a sliding scale where higher accuracy testing consumes greater computational resources. As such, the trade-off between proactive and retroactive error handling needs to be carefully analysed per application. 
Future work should investigate the transferability of the approach across diverse environments and robotic platforms to establish generalizability. 

Ethics, safety, and governance considerations remain paramount; we align with and adhere to ROSA's ``Ethics for Embodied Agents'' framework~\cite{royce2024enabling} in the development process. Several technical improvements could enhance agent performance, including strategies to improve tool invocation reliability, and mitigate hallucination and reduce latency.  %
Particularly promising is the development of ROS-specific embedding models through fine-tuning existing transformer architectures to improve vector database performance. 

\section*{Acknowledgment}
The authorship team would like to acknowledge the vision, support and guidance of the IEEE Industrial Electronics Society in conducting the Generative AI Hackathon under the leadership of Daswin De Silva and Lakshitha Gunasekara. 
We also gratefully acknowledge the mentorship and valuable insights provided by Rashmika Nawaratne, Leimin Tian, Brendan Tidd, and Dana Kuli{\'c}, whose contributions were instrumental to the success of this initiative.
\bibliographystyle{IEEEtran}
\bibliography{references} %

\end{document}